\documentclass[preprint,5p,authoryear]{elsarticle}

\usepackage{amsmath,amssymb,amsfonts}
\usepackage{algpseudocode,algorithm}
\usepackage{subcaption}
\usepackage{graphicx}
\usepackage{color}

\journal{Results in Control and Optimization}

\begin{document}

\sloppy

\begin{frontmatter}



\title{Reward Bonuses with Gain Scheduling Inspired by Iterative Deepening Search}


\author{Taisuke Kobayashi\corref{cor}}
\ead{kobayashi@nii.ac.jp}
\ead[url]{http://kbys\_t.gitlab.io/en/}

\cortext[cor]{Corresponding author}

\address{Principles of Informatics Research Division, National Institute of Informatics, Tokyo, Japan; and School of Multidisciplinary Sciences, Department of Informatics, The Graduate University for Advanced Studies (SOKENDAI), Kanagawa, Japan}

\begin{abstract}

This paper introduces a novel method of adding intrinsic bonuses to task-oriented reward function in order to efficiently facilitate reinforcement learning search.
While various bonuses have been designed to date, they are analogous to the depth-first and breadth-first search algorithms in graph theory.
This paper, therefore, first designs two bonuses for each of them.
Then, a heuristic gain scheduling is applied to the designed bonuses, inspired by the iterative deepening search, which is known to inherit the advantages of the two search algorithms.
The proposed method is expected to allow agent to efficiently reach the best solution in deeper states by gradually exploring unknown states.
In three locomotion tasks with dense rewards and three simple tasks with sparse rewards, it is shown that the two types of bonuses contribute to the performance improvement of the different tasks complementarily.
In addition, by combining them with the proposed gain scheduling, all tasks can be accomplished with high performance.

\end{abstract}


\end{frontmatter}

\section{Introduction}

Reinforcement learning (RL) is a framework for optimizing an agent's action policy by relying on rewards obtained through interactions with an unknown environment~\citep{sutton2018reinforcement}, and has attracted a great deal of attention in the last decade.
Since RL can be regarded as a kind of data-driven control technology, it can be applied to advanced robot controllers required to resolve tasks in the environment that are difficult to model.
For example, several studies have reported the applications for physical human robot interaction~\citep{modares2015optimized}, manipulation of deformable objects~\citep{tsurumine2019deep}, and so on.

However, RL suffer from its poor learning efficiency due to the learning process with repeated trial-and-error interactions between the agent and the environment.
Since this is a major obstacle in practical use, various approaches have been taken.
For example, a sim-to-real technique~\citep{tan2018sim,matas2018sim,rudin2022learning} accumulates numerous experiences in parallel via a simulated environment, thus completing learning without running the agent in the real environment.
Although this approach does not resolve the inefficiency of the learning process itself, recent improvements in parallel processing performance using GPUs have made it possible to obtain robust controllers quickly.
Alternatively, model-based RL~\citep{chua2018deep,okada2020planet}, which actively exploits future predictions needed for RL by learning the unknown environment as a simulator, so-called world model~\citep{ha2018world}, is known empirically to improve learning efficiency.
It has also been reported that the efficient search performance can be enhanced by modeling the action policy as student-t distribution~\citep{kobayashi2019student} or more complex one~\citep{ward2019improving}, so that stochastic actions are more diverse.

This study aims to design bonuses to rewards, inspired by intrinsic motivation, which is one of the studies for improving the search ability.
This intrinsic motivation is considered to be the driving force for us to gain new experiences~\citep{cameron1994reinforcement,chentanez2004intrinsically}.
In fact, since rewards are the index of optimization in RL, it is natural to modify the learning process by acting on them, and various utilities can be expected if appropriate bonuses are given.
For example, using the policy entropy as a bonus promotes stochastic search, and can achieve good performance in finding global solutions while suppressing over-learning~\citep{haarnoja2018soft,choy2020sparse}.
If the rewards are designed sparsely ---non-zero rewards are given only when certain conditions are satisfied--- a bonus that highly values unknown states is necessarily given to encourage active search, since the direction of search is unknown only with the task-oriented sparse rewards~\citep{bellemare2016unifying,bougie2021fast,andrychowicz2017hindsight}.
Although not for promoting search, it has also been reported that generalization performance and learning stability can also be expected by punishing undesirable experiences~\citep{berseth2019smirl,parisi2019td}.
Since various bonuses are designed as intrinsic motivation from RL elements, please refer to the survey paper~\citep{aubret2019survey} for details.
Note that, although the analytical optimal policy may change depending on the added bonus in terms of reward shaping~\citep{ng1999policy}, this paper does not discuss this point due to emphasizing practicality rather than analysis and the fact that this issue can be avoided if the bonus converges to a constant value eventually.

Thus, the bonuses to rewards can facilitate the search towards unknown new situations.
Based on graph theory~\citep{west2001introduction}, this search tendency can be broadly classified into two types: depth-first search (DFS) and breadth-first search (BFS).
In general, BFS can find global solutions, but it requires a great deal of experiences for the task with a large state space because it gradually deepens the search from around the initial state while always maximizing the search width.
In addition, the task with sparse rewards, where there is no driving force to deepen the search, may never reach new situations.
On the other hand, DFS is not guaranteed to find global solutions, but it would be efficient in finding local solutions.
In addition, DFS can be an additional driving force to deepen the search towards unknown situations.
Of course, the bonuses can be strictly classified as either DFS or BFS, and yield exactly the same properties as in graph theory.
Even though, it can be easily imagined from this analogy that a single type of bonuses will fail at the tasks that cannot be accomplished with its properties.

\begin{figure}[tb]
    \centering
    \includegraphics[keepaspectratio=true,width=0.96\linewidth]{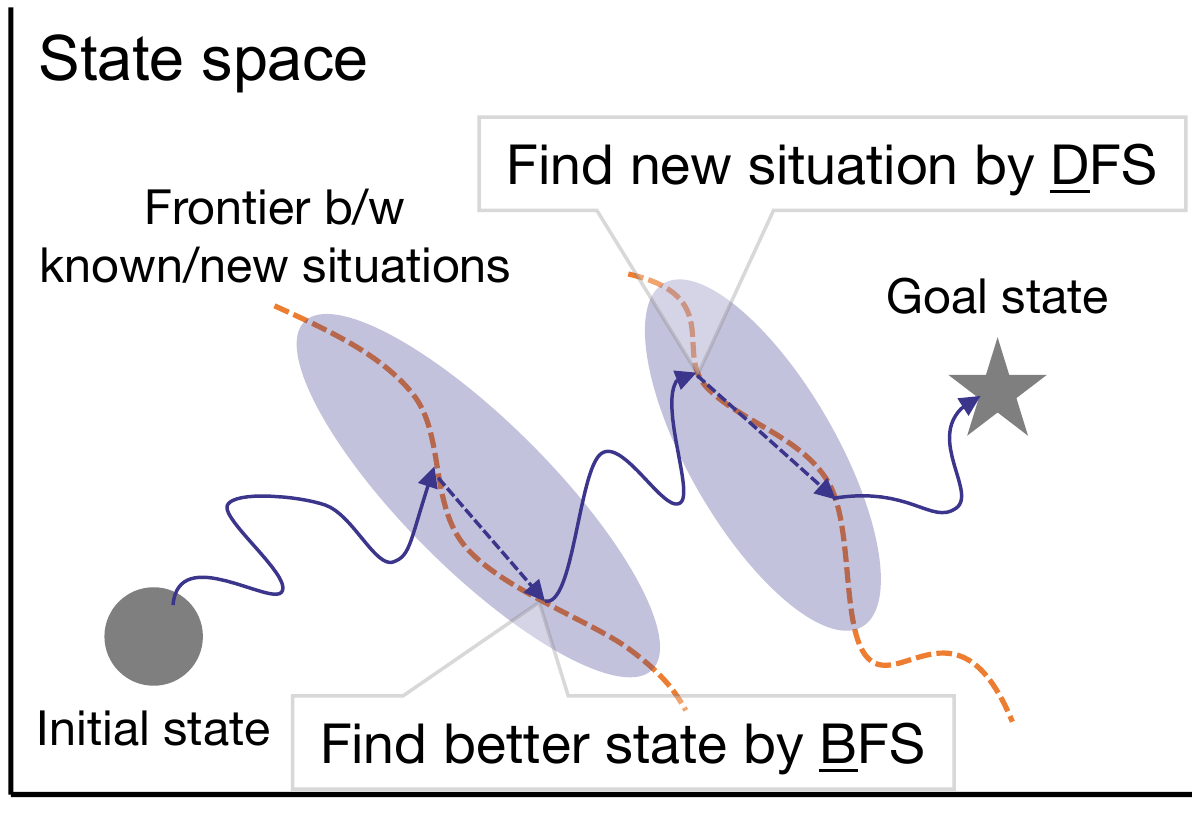}
    \caption{Concept of proposal}
    \label{fig:gain_scheduling}
\end{figure}

Therefore, two types of bonuses that can be broadly classified as DFS and BFS respectively are first designed.
Afterwards, inspired by iterative deepening search (IDS)~\citep{korf1985depth}, which appropriately combines DFS and BFS in graph theory, these two DFS- and BFS-like bonuses are combined with a well-balanced gain scheduling.
The proposed method allows the DFS-like bonus to steadily deepen the search towards new situations while the BFS-like bonus broadens the search range at the frontier to find better solutions (see Fig.~\ref{fig:gain_scheduling}).

Specifically, as DFS-like bonus, \textit{value disagreement} for ensemble-learned value functions is considered with reference to the disagreement proposed in the previous study~\citep{pathak2019self}.
On the other hand, BFS-like bonus is formulated as \textit{self-imitation} so as not to forget how to acquire important experiences in the past, while its derivation is differently given from the previous study~\citep{oh2018self}.
Then, IDS-inspired gain scheduling encourages deep dives into new situations in favor of DFS-like bonus if the policy and value functions are estimated to be stagnant; otherwise, it broadens the search around the current situation by selecting BFS-like bonus.
To achieve the right balance in this gain scheduling, automatic optimization of the relevant hyperparameters is also provided.

The properties of the designed bonuses are verified numerically on three locomotion tasks with dense rewards~\citep{coumans2016pybullet} and three simple tasks with sparse rewards~\citep{tunyasuvunakool2020dm_control}.
The results show that both bonuses tend to facilitate the search efficiently and yield the better performance than RL without them, but that the tasks suitable for them are complementary and different.
Thanks to IDS-inspired gain scheduling, more generalized task accomplishment performance is finally achieved by properly inheriting the properties of both types of bonuses.

The contributions of this paper are three folds:
\begin{enumerate}
    \item A new perspective that bonuses to rewards as intrinsic motivation can be classified as DFS or BFS from the perspective of graph theory.
    \item Design of two types of bonuses based on DFS and BFS respectively.
    \item Development of a heuristic for scheduling the gains of the two bonuses to achieve IDS-like behavior.
\end{enumerate}
While there would be room for improvement in any of the designs, to the best of my knowledge, this is the first study to classify and integrate bonuses with this concept, and its usefulness has been reliably verified by numerical simulations.

\section{Preliminaries}

\subsection{Reinforcement learning}

The basic problem statement of RL~\citep{sutton2018reinforcement} is briefly introduced at first.
Note that the proposed method can be applied to any algorithms.

RL has a trainable agent and an unknown environment in its framework of Markov decision process.
At the current state $s$, the agent takes action $a$ according to a state-dependent trainable policy, $\pi(a \mid s; \theta_{\pi})$, with its parameters $\theta_{\pi}$.
With $a$, the environment updates $s$ to the next one $s^\prime$ by its state transition probability $p_e(s^\prime \mid s, a)$.
At the same time, the environment evaluates the above process as reward $r = r(s, a, s^\prime)$.
Note that the initial state is given from the own probability $p_0(s)$.
In recent RL algorithms, the numerous amount of experience gained by iterating the above process is stored in a buffer and replayed as needed at learning~\citep{schaul2015prioritized}.

The main objective of RL is to maximize the sum of rewards over future with the discount factor $\gamma$ (i.e. return) by optimizing $\pi$.
However, future information must be predicted, RL often defines a trainable (state) value function, $V(s; \theta_{V})$, with its parameters $\theta_{V}$, in order to predict the return from the current state with $\pi$.
Finally, RL optimizes $\pi$ to maximize $V$ while increasing the accuracy of $V$.
Although various algorithms have been proposed for this problem statement, this paper adopts an implementation based on a latest actor-critic algorithm~\citep{kobayashi2022consolidated}.

\subsection{Ensemble learning}

\begin{figure}[tb]
    \centering
    \includegraphics[keepaspectratio=true,width=0.96\linewidth]{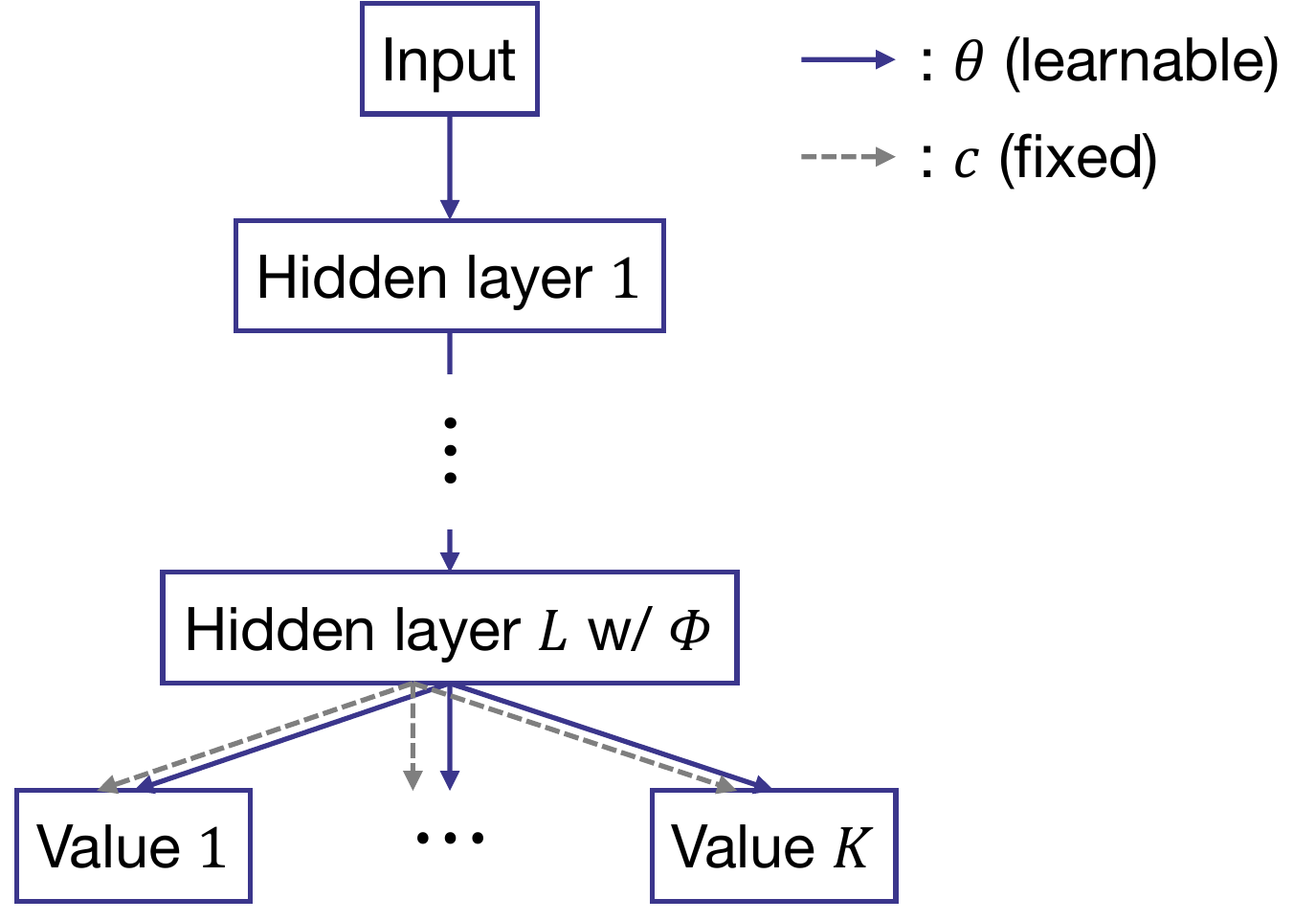}
    \caption{Ensembled network architecture}
    \label{fig:net_ensemble}
\end{figure}

In the above RL, the temporal difference (TD) error $\delta$ is commonly used to learn $V$ (and $\pi$).
\begin{align}
    \delta &= r + \gamma V(s^\prime, \theta_{V}) - V(s, \theta_{V})
    \label{eq:td_error}\\
    \theta_{\pi,V} &\gets \theta_{\pi,V} + \alpha \delta g_{\pi,V}
    \label{eq:rl_update}
\end{align}
where $\alpha > 0$ denotes the learning rate and $g_{\pi} = \nabla_{\theta_{\pi}} \ln \pi(a \mid s; \theta_{\pi})$ and $g_{V} = \nabla_{\theta_{V}} V(s, \theta_{V})$, respectively.
However, since this method is learning with pseudo-supervised signals derived from the structural properties of $V$, its accuracy is inevitably insufficient and over-fitting is likely to occur.
To alleviate these problems, ensemble learning of $V$ is often employed~\citep{osband2016deep,osband2018randomized}.

Specifically, $V(s; \theta_{V})$ is first expanded as follows:
\begin{align}
    V(s; \theta_{V}) = \{V_k(s; \theta_{V_k}) + \beta C_k(s; c_k)\}_{k=1}^K
    \label{eq:value_ensemble}
\end{align}
where $K$ denotes the number of ensembles.
$C_k$ with $c_k$ the fixed parameters is a differentially initialized prior that can be multiplied by $\beta \geq 0$ and added to $V_k$ to improve the generalization performance and make it easier for each element to output different values for untrained states~\citep{osband2018randomized}.
The consensus value of $V$, $\bar{V}$, is used as an estimate for learning.
In many cases, the mean operation is used for the consensus, but since the effect of noise remains when $K$ is not enough, the median operation is employed in this paper (see \ref{app:median}).

In terms of the computational and memory costs, completely different $\theta_{V_k}$ would waste them.
Instead, the trainable feature $\Phi(s; \theta_{\Phi})$ (with $\theta_{\Phi}$ the trainable parameters) is shared, and then it is transformed to $V_k$ linearly using $\theta_{V_k}$~\citep{osband2016deep}.
Similarly, $C_k$ shares $\Phi$ and transforms it using $c_k$.
\begin{align}
    V_k(s; \theta_{V_k}) = (\theta_{V_k} + \beta c_k)^\top \Phi(s; \theta_{\Phi})
    \label{eq:value_ensemble2}
\end{align}
This can be easily implemented using neural networks by sharing them until the final hidden layer and outputting $K$ values in the output layer, as shown in Fig.~\ref{fig:net_ensemble}.
Note that $\theta_{V} = \{\theta_{V_k}\}_{k=1}^K \cup \theta_{\Phi}$ in this case.

\section{Proposal}

\subsection{Overview}

In RL, more efficient search is needed to learn the value of various states in an unknown environment.
Many studies have been conducted to add bonuses to task-oriented rewards as a means of promoting the search~\citep{aubret2019survey}.
This paper attempts to design appropriate bonuses following these studies.

To this end, it is noticeable that, according to graph theory, search can be broadly classified into DFS and BFS~\citep{west2001introduction}, even in the case of adding bonuses.
Each of these search approaches has its own properties, and since the memory efficiency aspect can be regarded as the sample efficiency in RL, the tasks for which they are effective would be different.
Therefore, IDS, which combines these two types of search approaches appropriately~\citep{korf1985depth}, is desired to be reproduced by a gain scheduling of two types of bonuses to search for states that further away from the initial state little by little without missing any states.

Specifically, two bonuses, $r_d$ for DFS and $r_b$ for BFS respectively, are added to the task-oriented reward $r$ with an adaptive gain scheduler $\zeta \in [0, 1]$ as follows:
\begin{align}
    r \gets r + \lambda \left \{ \zeta r_d + (1 - \zeta) r_b  \right \}
    \label{eq:bonus_base}
\end{align}
where $\lambda \geq 0$ denotes the base gain for the bonuses.
According to this framework, $r_d$, $r_b$, and $\zeta$ are appropriately designed, as introduced in the next sections.

\subsection{Value disagreement as depth first search}

\begin{figure}[tb]
    \centering
    \includegraphics[keepaspectratio=true,width=0.96\linewidth]{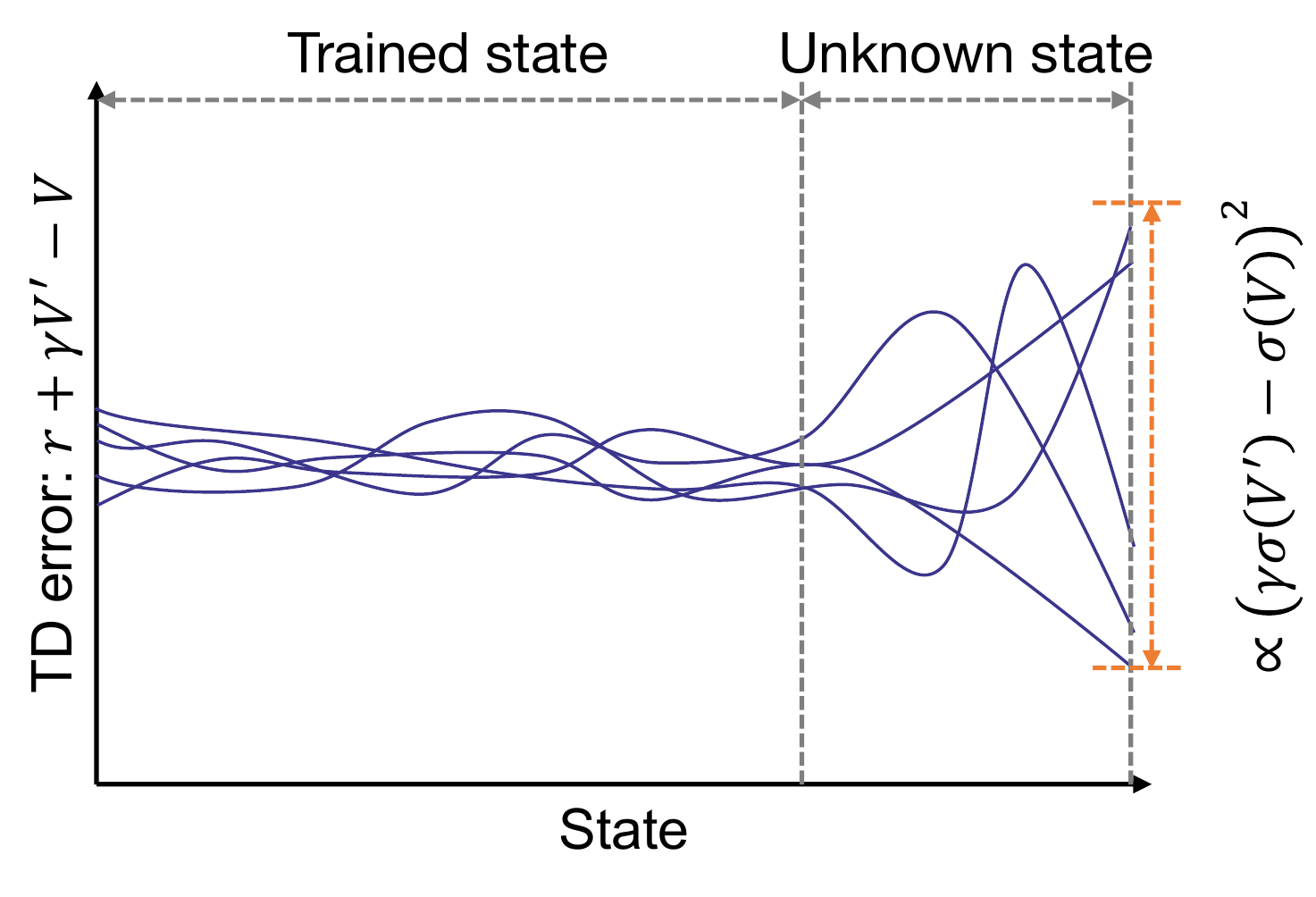}
    \caption{Metric of value disagreement}
    \label{fig:value_disagreement}
\end{figure}

In the previous work, \textit{disagreement}~\citep{pathak2019self} has been proposed and designed as a bonus that suggests states that have not yet been explored.
That is, by learning an ensemble of state transition probability models $\{q_e^k\}_{k=1}^K \to p_e$ from the stored experiences, the variance of states predicted by these multiple components with a given pair of state and action (a.k.a. the disagreement) can be calculated:
if the disagreement is large, the models suggest that the given pair of state and action would be inexperienced enough and should be preferentially searched.
Such a disagreement-based bonus can be interpreted as a kind of DFS that leads to a deeper search into unknown states.
This paper therefore designs a DFS-like bonus that follows this idea, disagreement.

Specifically, since additionally learning $\{q_e^k\}_{k=1}^K$ is wasteful if they are utilized only for the disagreement computation, the disagreement of the value function $V$ is considered instead.
Since the ensemble of $\{V_k\}_{k=1}^K$ is already introduced for stabilizing TD learning, as defined in eq.~\eqref{eq:value_ensemble2}, they can be reused as-is.
However, the median operation for computing the consensus value is better to suppress the effects of noise rather than the mean operation, as discussed in \ref{app:median}.
According to this fact, the median absolute deviation would be better as the disagreement, instead of the variance (or, the standard deviation).
\begin{align}
    \sigma(V(s; \theta_{V})) &= \mathrm{Median}_{k=1,\ldots,K}(|V_k(s; \theta_{V_k}) - \bar{V}|)
    \label{eq:value_mad} \\
    \bar{V} &= \mathrm{Median}_{k=1,\ldots,K}(V_k(s; \theta_{V_k}))
    \nonumber
\end{align}
Using this, even if outliers are mixed in the $K$ components, their effects can be mitigated, outputting the stable disagreement.
Note that, instead of $V$, the action value function $Q$ would be able to compute the disagreement as well.

Another concern is that the variance is generally divided into epistemic and aleatoric factors, and even if the epistemic one is sufficiently eliminated after sufficient learning, the aleatoric one will still remain.
Hence, if $\sigma(V(s; \theta_{V}))$ is utilized as-is, the search will favor states with large aleatoric uncertainty, i.e. states where uncertain transitions are likely to occur.
In that case, the sample efficiency may be deteriorated.
Therefore, the following index is introduced as the disagreement with reference to the literature~\citep{parisi2019td}, which prioritizes the experience with large variance of TD error $\delta$ (see Fig.~\ref{fig:value_disagreement}).
\begin{align}
    \mathrm{Var}_{k=1,\ldots,K}[\delta_k] &= \mathbb{E}_{k=1,\ldots,K}[\delta_k^2]
    - \mathbb{E}_{k=1,\ldots,K}[\delta_k]^2
    \nonumber \\
    &= \mathbb{E}_{k=1,\ldots,K}[\{ \gamma V_k(s^\prime; \theta_{V_k}) - V_k(s; \theta_{V_k}) \}^2]
    \nonumber \\
    &- \mathbb{E}_{k=1,\ldots,K}[ \gamma V_k(s^\prime; \theta_{V_k}) - V_k(s; \theta_{V_k}) ]^2
    \nonumber \\
    &= \gamma^2 \mathrm{Var}_{k=1,\ldots,K}[V_k(s^\prime; \theta_{V_k})]
    \nonumber \\
    &+ \mathrm{Var}_{k=1,\ldots,K}[V_k(s; \theta_{V_k})]
    \nonumber \\
    &- 2 \gamma \mathrm{Cov}_{k=1,\ldots,K}[V_k(s^\prime; \theta_{V_k}), V_k(s; \theta_{V_k})]
    \nonumber \\
    &\propto \left\{ \gamma \sigma(V(s^\prime; \theta_{V})) - \sigma(V(s; \theta_{V})) \right \}^2
\end{align}
where the covariance is approximated as square root of the product of the corresponding variances, and the variance is replaced by square of the median absolute deviation.
With this definition, the aleatoric uncertainty can be mitigated due to the relative evaluation of the scales in the current state and the next state.

However, even when both states are inexperienced and the epistemic uncertainties for both are also large, the relative evaluation may cancel not only the aleatoric uncertainty but also the epistemic one, so there is room for consideration of how much to make the evaluation relative.
Therefore, a DFS-like bonus based on the above is designed as follows:
\begin{align}
    r_d = |\gamma \sigma(V(s^\prime; \theta_{V})) - \eta_d \sigma(V(s; \theta_{V}))|^{\nu_d}
    \label{eq:bonus_dfs}
\end{align}
where $\eta_d \in [0, 1]$ denotes the relative evaluation ratio, and by increasing $\nu_d > 0$, the bonus can be easily changed.
Note that the introduction of $\eta_d$ is a compromise with the bonus of prioritizing experience with high uncertainty in the pseudo-supervised signal, as in the literature~\citep{flennerhag2020temporal}.

\subsection{Self-imitation as breadth first search}

\begin{figure}[tb]
    \centering
    \includegraphics[keepaspectratio=true,width=0.96\linewidth]{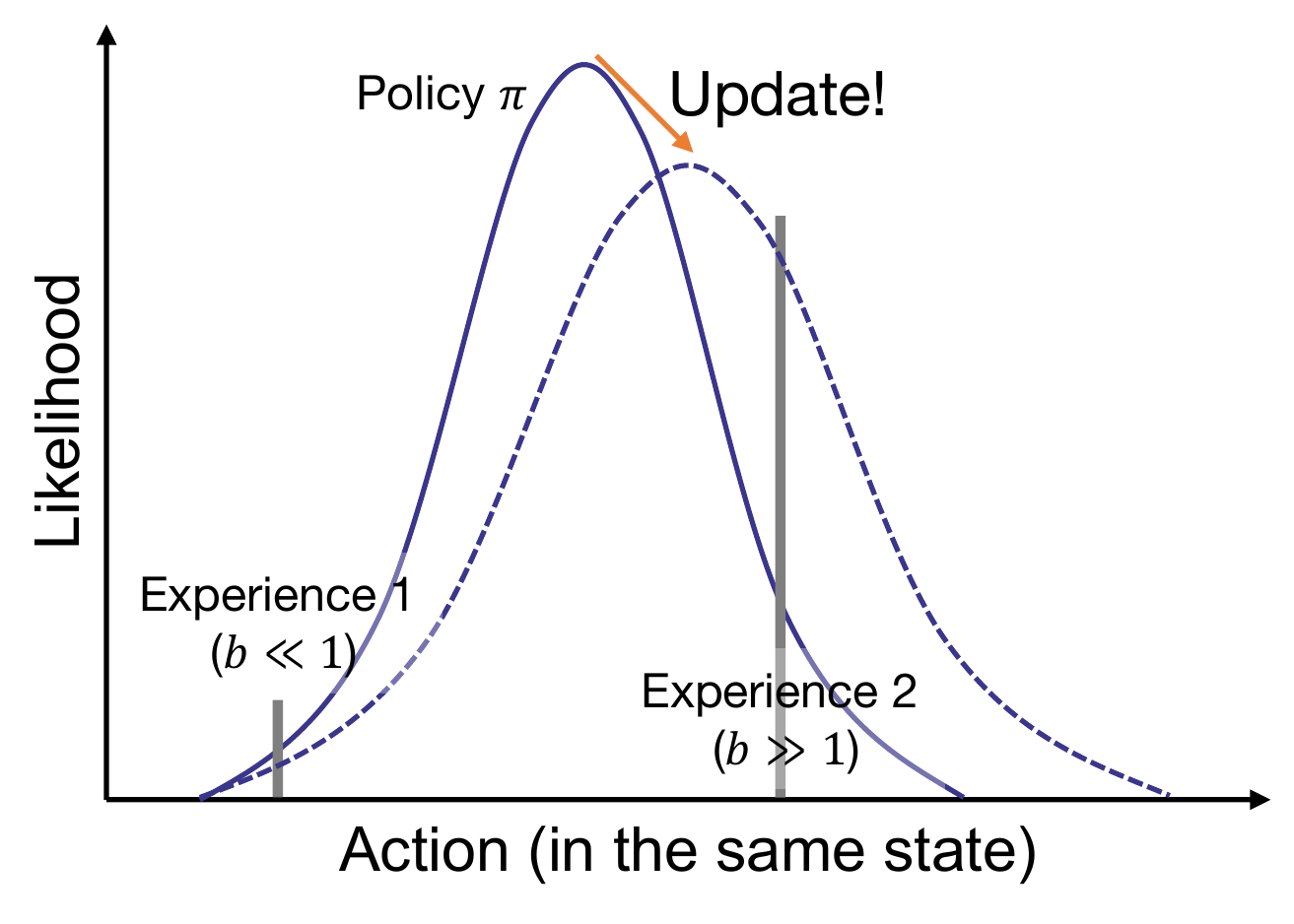}
    \caption{Example of self-imitation}
    \label{fig:self_imitation}
\end{figure}

A well-known bonus for facilitating search is the policy entropy~\citep{haarnoja2018soft}.
By increasing the randomness of the actions generated by the policy $\pi$, more diverge experiences can be gained expectedly.
With broadening the range of search, this can be interpreted as a kind of BFS.
This paper therefore designs a similar bonus.

Specifically, in addition to the update rule based on the policy-gradient method in eq.~\eqref{eq:rl_update}, an auxiliary problem of minimizing the deviation from the policy generating the experiences in the past, $b$.
In other words, the search width is maintained and/or broadened so as to visit states experienced in the past by self-imitation~\citep{oh2018self}.
However, it is not necessary to treat all experiences as equivalent, and it may be desirable to preferentially imitate more common transitions, rather than ones occurred by chance.
For such a requirement, the following self-imitation is derived from minimization of Tsallis divergence~\citep{gil2013renyi}.
\begin{align}
    & \mathbb{E}_{s \sim p_e} \left[ \mathrm{KL}_q(b(a \mid s) || \pi(a \mid s; \theta_{\pi})) \right]
    \nonumber \\
    =& - \mathbb{E}_{s \sim p_e, a \sim b} \left[ \ln_q \frac{\pi(a \mid s; \theta_{\pi})}{b(a \mid s)} \right]
    \nonumber \\
    =& - \mathbb{E}_{s \sim p_e, a \sim b} \left[ b^{q-1}(a \mid s) \ln_q \pi(a \mid s; \theta_{\pi}) \right]
    - \mathbb{E}_{s \sim p_e} \left[ H_q(b) \right]
    \nonumber \\
    \propto& - \mathbb{E}_{(s, a) \sim D} \left[ b^{q-1}(a \mid s) \ln_q \pi(a \mid s; \theta_{\pi}) \right]
    = \mathcal{L}_{si}(\theta_{\pi})
    \label{eq:loss_qbc_kl}
\end{align}
where $q \in \mathbb{R}$ denotes the hyperparameter for extending natural logarithm to $q$-logarithm.
The gradient of this loss function is given as follows:
\begin{align}
    \nabla_{\theta_{\pi}} \mathcal{L}_{si}(\theta_{\pi})
    &= - \mathbb{E}_{(s, a) \sim D} \left[ \rho \ln \pi(a \mid s; \theta_{\pi}) \right]
    \nonumber \\
    \rho &= \exp\{(1 - q)(\ln\pi(a \mid s; \theta_{\pi}) - \ln b(a \mid s))\}
    \label{eq:grad_ratio_kl}
\end{align}
When minimizing this auxiliary problem simultaneously with eq.~\eqref{eq:rl_update}, the original policy-gradient is modified with $\delta + \rho$ the coefficient.
That is, $\rho$ can be regarded as a bonus added to the task-oriented reward.

According to this interpretation, a more appropriate bonus is designed based on $\rho$.
First, as for $q$, it is desirable to set $q > 1$ to correspond to the policy entropy mentioned at the beginning of this section.
In this way, since $(1 - q) = - \nu_b < 0$, a term corresponding to the policy entropy $- \nu_b \ln \pi$, variation of which can be tuned by $\nu_b$, appears in the exponential function.
At the same time, the inclusion of $\nu_b \ln b$ prioritizes the experiences generated with higher likelihoods (see Fig.~\ref{fig:self_imitation}).
Note that, if this prioritization is performed excessively, it will ignore valuable experiences as well, and therefore, an additional adjustment is considered necessary.
In summary, a BFS-like bonus is designed as follows:
\begin{align}
    r_b = \exp \left \{ - \nu_b (\ln \pi(a \mid s; \theta_{\pi}) - \eta_b \ln b) \right \}
    \label{eq:bonus_bfs}
\end{align}
where $\eta_b \in [0, 1]$ is responsible for adjusting the degree of reflection of $\ln b$, which can be stored into the buffer together with the corresponding experience $(s, a, s^\prime, r)$.
Note that the introduction of $\eta_b$ is a compromise with the approach of considering imitation learning as a $q$-logarithm maximization problem where the term of $\ln b$ does not appear, as in the literature~\citep{kobayashi2021towards}.

\subsection{Gain scheduling as iterative deepening search}

In order to reproduce IDS that appropriately combines the properties of DFS and BFS, we design $\zeta$ in eq.~\eqref{eq:bonus_base}.
The key is to design a metric that determines whether the replayed experience is sufficiently learned and regarded to be stagnant.
At the stagnation, $r_d$ would be more effective than $r_b$ to aggressively find new states, otherwise, $r_b$ should be prioritized to broaden the search range (see Fig.~\ref{fig:gain_scheduling}).
In this way, IDS-like behavior, which gradually deepens the search while widening it to ensure the whole search, can be expected with this design.

Specifically, the learning stagnation is first defined by reusing the elements used to calculate $r_d$ and $r_b$.
That is, the search is considered to be stagnant at $\sigma(V(s^\prime; \theta_{V})) \simeq \sigma(V(s; \theta_{V}))$ and at $\pi(a \mid s; \theta_{\pi}) \simeq b$.
A metric, $m_{\kappa}(x, y) \in (0, 1)$ with $x, y, \kappa > 0$, is therefore defined based on the inequality of arithmetic and geometric means.
\begin{align}
    m_{\kappa}(x, y) = \left \{ 1 - \frac{|x - y|^\kappa}{(x + y)^\kappa} \right \}^{\frac{1}{\kappa}}
\end{align}
Here, $\kappa$ determines the shape of $m$: if $\kappa \ll 1$, $m_{\kappa}$ outputs $0$ except for $x \simeq y$; and if $\kappa \gg 1$, $m_{\kappa}$ outputs $1$ mostly.
Using this stagnation metric, $\zeta$ is defined as follows:
\begin{align}
    \zeta = \sqrt{m_{\kappa_d}(\sigma(V(s^\prime; \theta_{V})), \sigma(V(s; \theta_{V}))) m_{\kappa_b}(\pi(a \mid s; \theta_{\pi}), b)}
    \label{eq:gain_schedule}
\end{align}
where $\kappa_{d,b} > 0$ denote the hyperparameters, which will be optimized as below.

If $\kappa_{d,b}$ is poorly tuned, only either of $r_{d,b}$ is activated, resulting in an unbalanced search.
Therefore, the following minimization problem and its gradient is considered for $\zeta \to 1/2$.
\begin{align}
    &\quad \min_{\kappa_{d,b}} \left| \zeta - \frac{1}{2} \right|
    \nonumber \\
    \nabla_{\kappa_{d,b}} \left| \zeta - \frac{1}{2} \right| &= \mathrm{sign}\left( \zeta - \frac{1}{2} \right) \frac{m_{\kappa_{b,d}}}{\zeta} \nabla_{\kappa_{d,b}} m_{\kappa_{d,b}}
    \\
    \nabla_{\kappa} m_{\kappa} &= - \frac{m_{\kappa}}{\kappa} \left \{ \ln m_{\kappa} + \frac{1 - m_{\kappa}^{\kappa}}{\kappa m_{\kappa}^{\kappa}} \ln(1 - m_{\kappa}^{\kappa}) \right \}
    \nonumber
\end{align}
With the derived gradient, $\kappa_{d,b}$ can be optimized under the consideration of $\kappa_{d,b} > 0$.
That is, instead of the standard gradient descent, exponentiated gradient descent is employed as follows:
\begin{align}
    \kappa_{d,b} \gets \kappa_{d,b} \exp \left \{ - \alpha_{\kappa} \nabla_{\kappa_{d,b}} \left| \zeta - \frac{1}{2} \right| \right \}
\end{align}
where $\alpha_{\kappa} > 0$ denotes the learning rate.

\section{Simulations}

\subsection{Setup}

\begin{figure}[tb]
    \centering
    \includegraphics[keepaspectratio=true,width=0.96\linewidth]{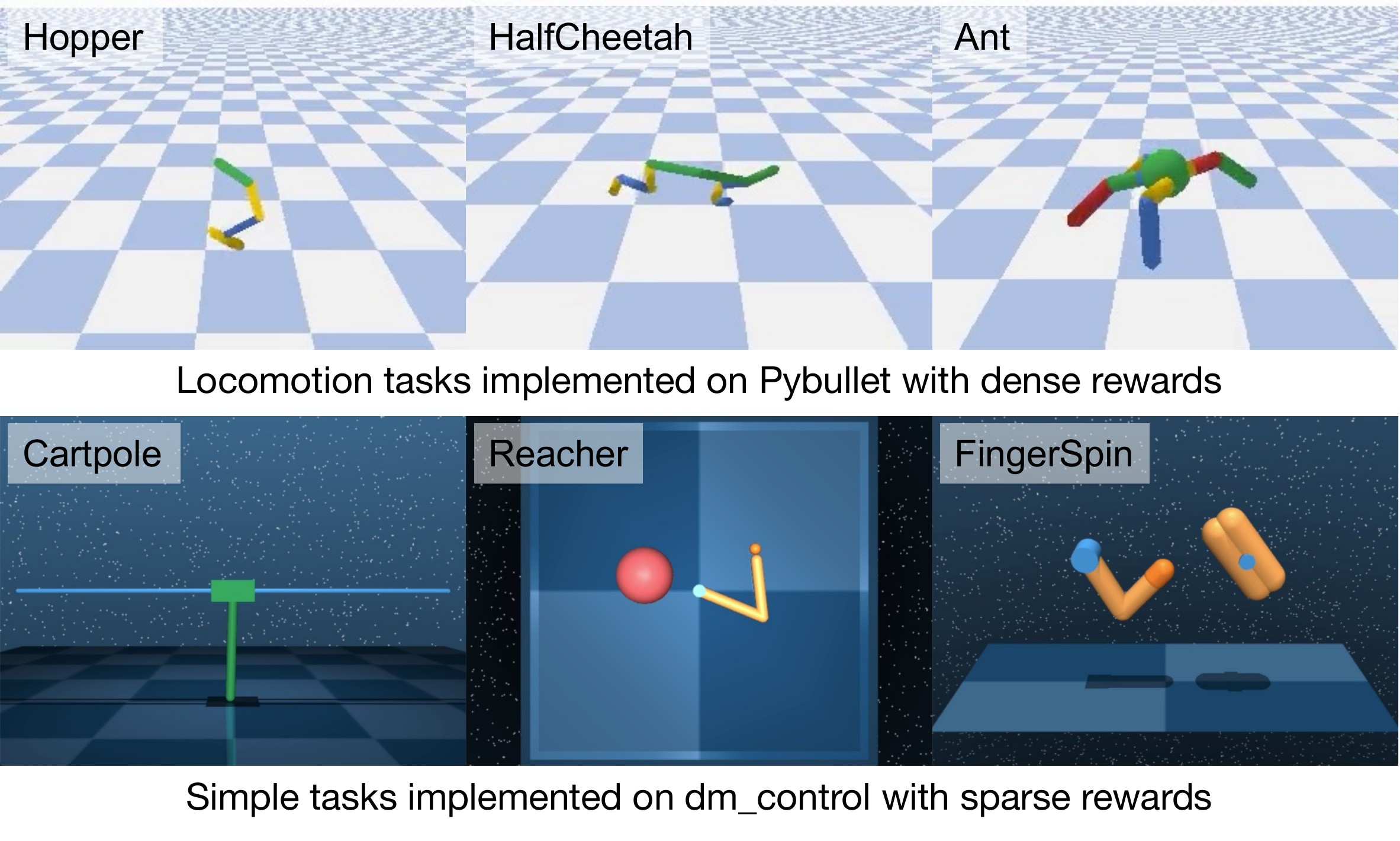}
    \caption{Snapshots of tasks conducted}
    \label{fig:snap_task}
\end{figure}

\begin{table}[tb]
    \caption{Hyperparameters for the proposed method}
    \label{tab:param}
    \centering
    \begin{tabular}{ccc}
        \hline\hline
        Symbol & Meaning & Value
        \\
        \hline
        $\lambda$ & Basic scale of bonuses & $0.1$
        \\
        $\eta_d$ & Relative ratio in DFS-like bonus & $0.5$
        \\
        $\nu_d$ & Power in DFS-like bonus & $2$
        \\
        $\eta_b$ & Relative ratio in BFS-like bonus & $0.5$
        \\
        $\nu_b$ & Power in BFS-like bonus & $0.1$
        \\
        \hline
        $\alpha_{\kappa}$ & Learning rate for $\kappa_{d,b}$ & $0.1\alpha$
        \\
        $\kappa_{d,b}^{\mathrm{ini}}$ & Initial $\kappa_{d,b}$ & $1$
        \\
        \hline\hline
    \end{tabular}
\end{table}

For the statistical verification of the proposed method, the following simulations, as illustrated in Fig.~\ref{fig:snap_task}, are conducted:
\begin{itemize}
    \item \textit{HopperBulletEnv-v0} (Hopper)
    \item \textit{HalfCheetahBulletEnv-v0} (HalfCheetah)
    \item \textit{AntBulletEnv-v0} (Ant)
    \item \textit{CartpoleSwingupSparseDMC-v0} (Cartpole)
    \item \textit{ReacherEasyDMC-v0} (Reacher)
    \item \textit{FingerSpinDMC-v0} (FingerSpin)
\end{itemize}
The first three tasks are implemented on Pybullet~\citep{coumans2016pybullet}, and the remaining three tasks are provided by dm\_control~\citep{tunyasuvunakool2020dm_control}, where \texttt{frame\_skip} is set to 2.
The former tasks are with dense rewards, but due to the relatively large state-action space, the active search contributes significantly to control performance.
The latter tasks are relative simple, but sparse rewards are designed, requiring the active search.
To increase the difficulty of the tasks, Gaussian noise with a noise scale of $10^{-3}$ is added to observation every time.

As a baseline RL algorithm, the latest actor-critic method~\citep{kobayashi2022consolidated} is employed.
It is basically implemented as original (details are in \ref{app:impl}), except ensemble learning for the value function (see \ref{app:median}) and the proposed reward bonuses.
Hyperparamters introduced in the proposed method are roughly tuned empirically, as can be seen in Table~\ref{tab:param}.

The following four conditions are compared to verify the proposed method:
\begin{itemize}
    \item Vanilla without any bonuses ($\lambda=0$)
    \item DFS only with DFS-like bonus ($\zeta=1$)
    \item BFS only with BFS-like bonus ($\zeta=0$)
    \item Proposal with gain scheduling for two bonuses
\end{itemize}
In addition, two ablation studies are also tested: the mean of two bonuses with $\zeta=0.5$ (Mean); and the sum of two bonuses with $\zeta=0.5$ and $\lambda=0.2$ (Sum).
With one random seed (24 seeds in total), each task is tried to be accomplished by each method.
The sum of rewards is computed for each episode as a score (larger is better).
The trained policy is run 100 times for evaluating the final scores.

\subsection{Result}

\begin{figure*}[tb]
    \begin{subfigure}[b]{0.32\linewidth}
        \centering
        \includegraphics[keepaspectratio=true,width=\linewidth]{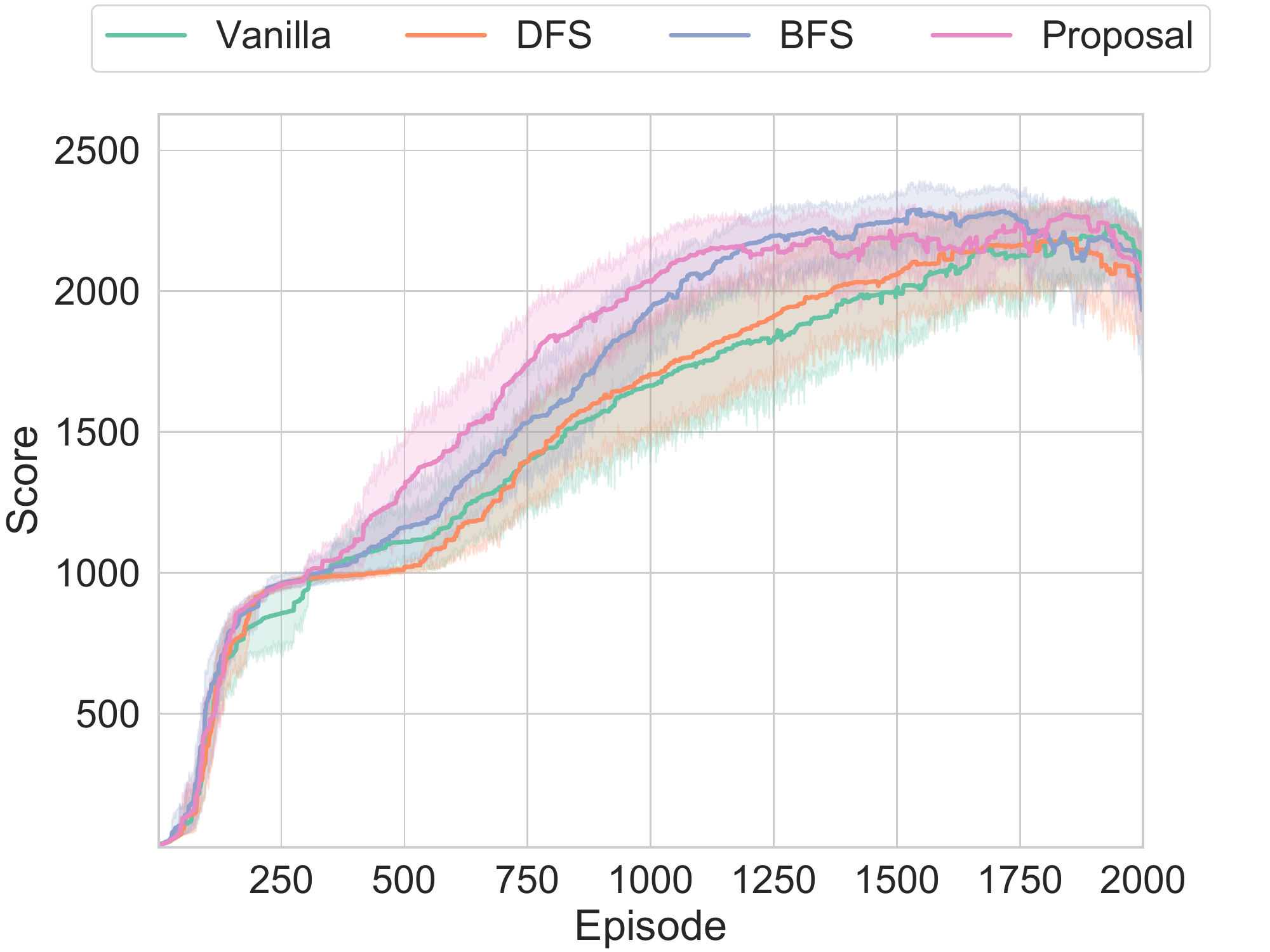}
        \subcaption{Hopper}
        \label{fig:result_score_Hopper}
    \end{subfigure}
    \begin{subfigure}[b]{0.32\linewidth}
        \centering
        \includegraphics[keepaspectratio=true,width=\linewidth]{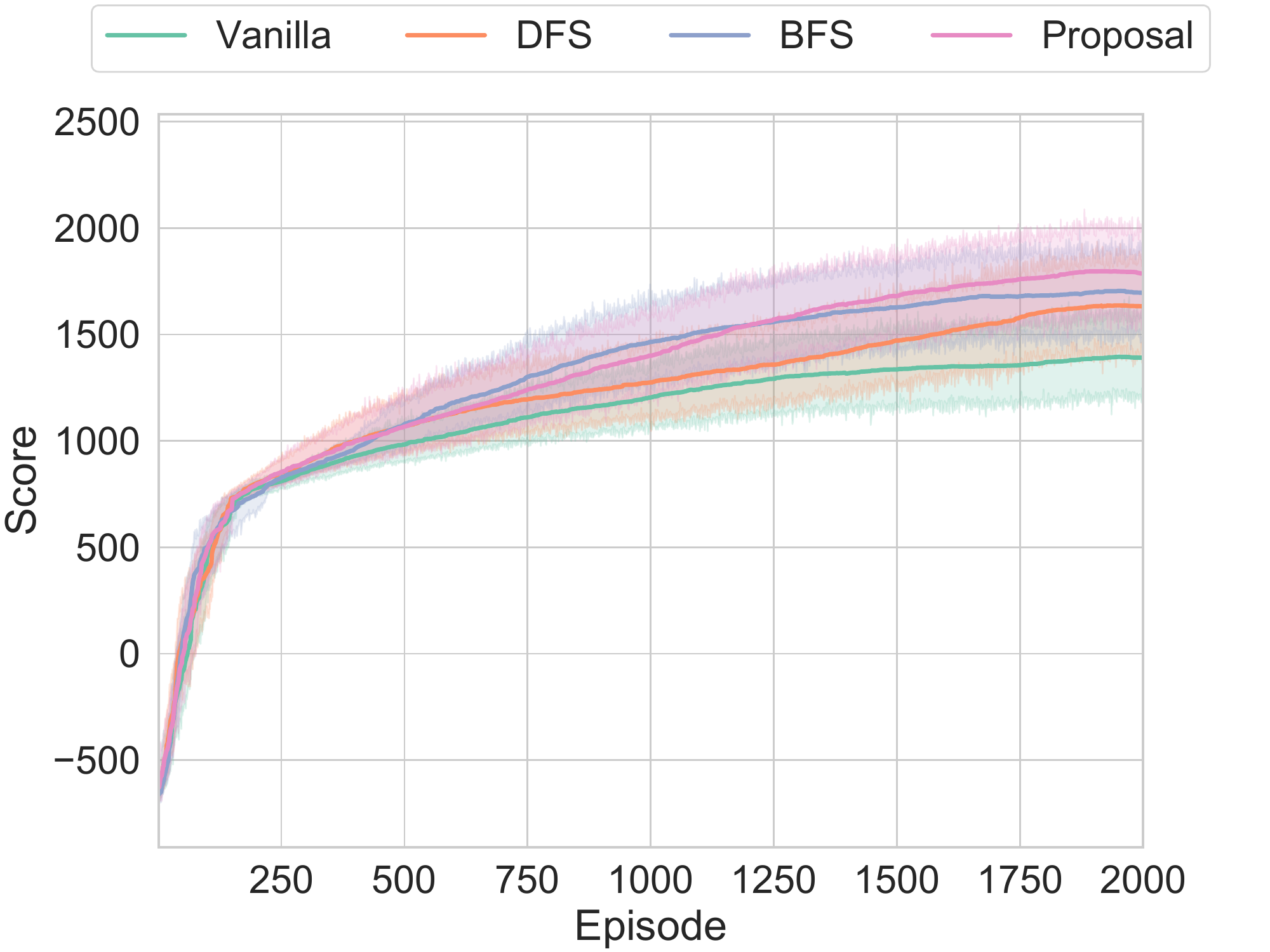}
        \subcaption{HalfCheetah}
        \label{fig:result_score_HalfCheetah}
    \end{subfigure}
    \begin{subfigure}[b]{0.32\linewidth}
        \centering
        \includegraphics[keepaspectratio=true,width=\linewidth]{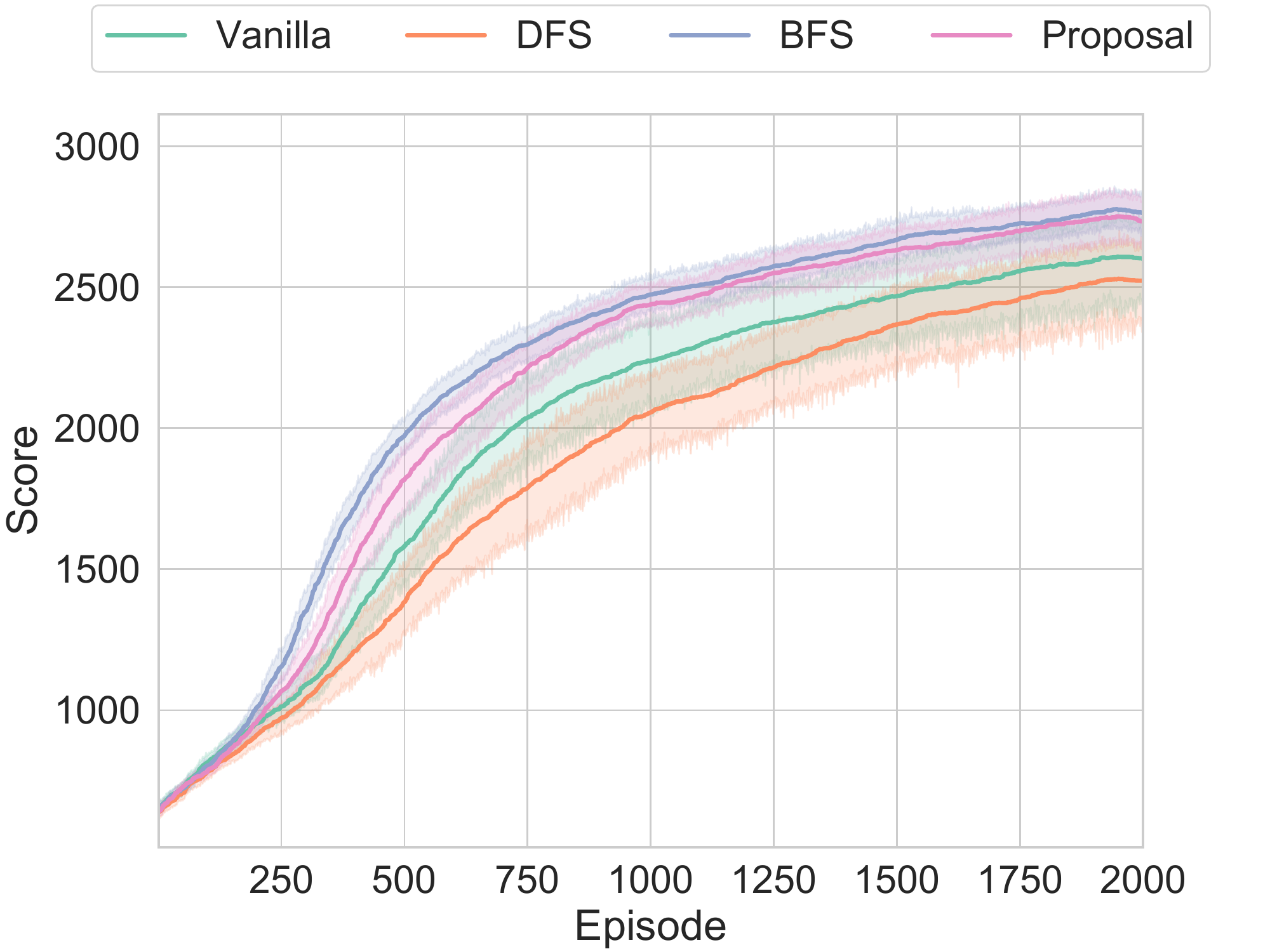}
        \subcaption{Ant}
        \label{fig:result_score_Ant}
    \end{subfigure}
    \begin{subfigure}[b]{0.32\linewidth}
        \centering
        \includegraphics[keepaspectratio=true,width=\linewidth]{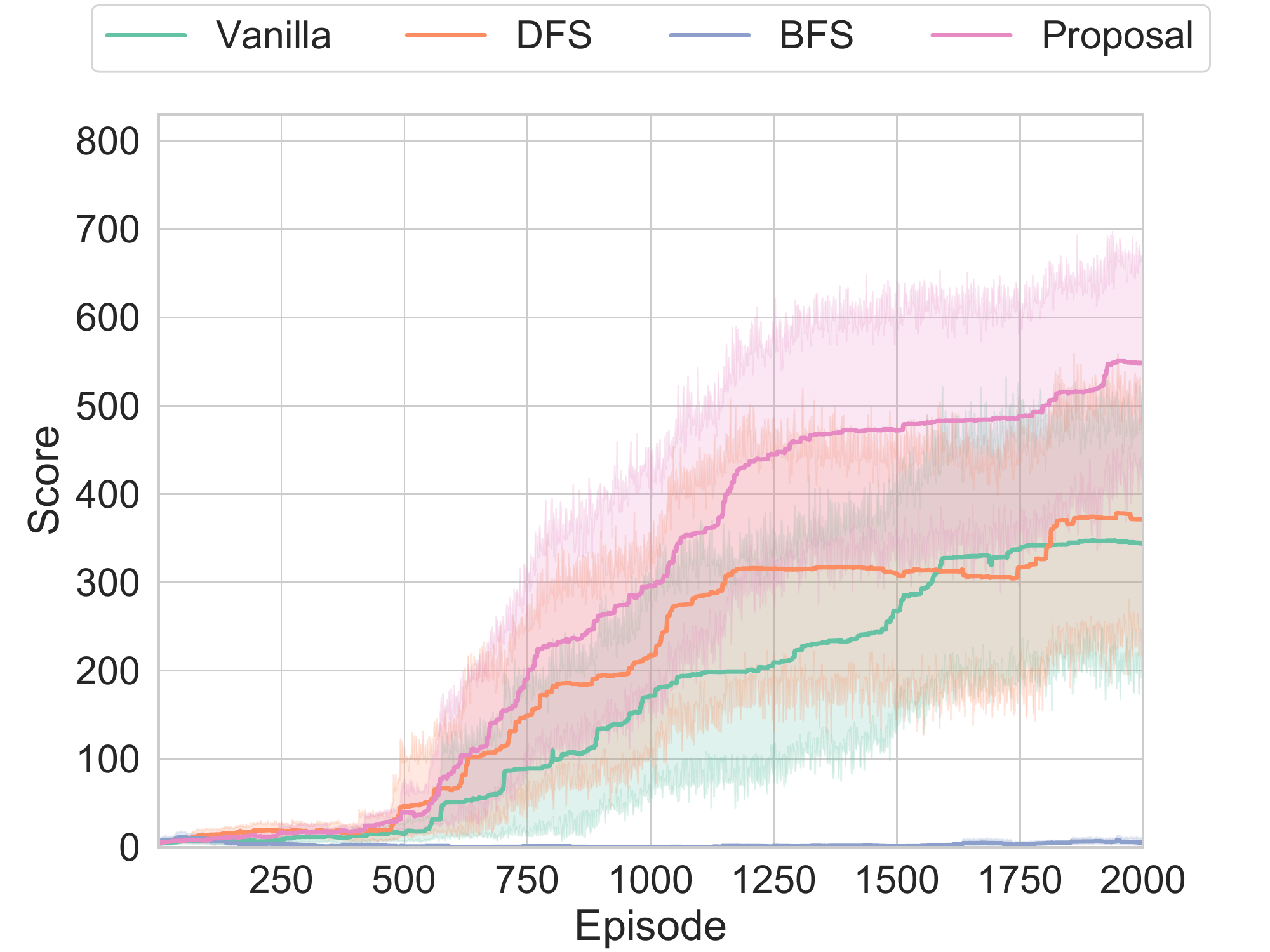}
        \subcaption{Cartpole}
        \label{fig:result_score_Cartpole}
    \end{subfigure}
    \begin{subfigure}[b]{0.32\linewidth}
        \centering
        \includegraphics[keepaspectratio=true,width=\linewidth]{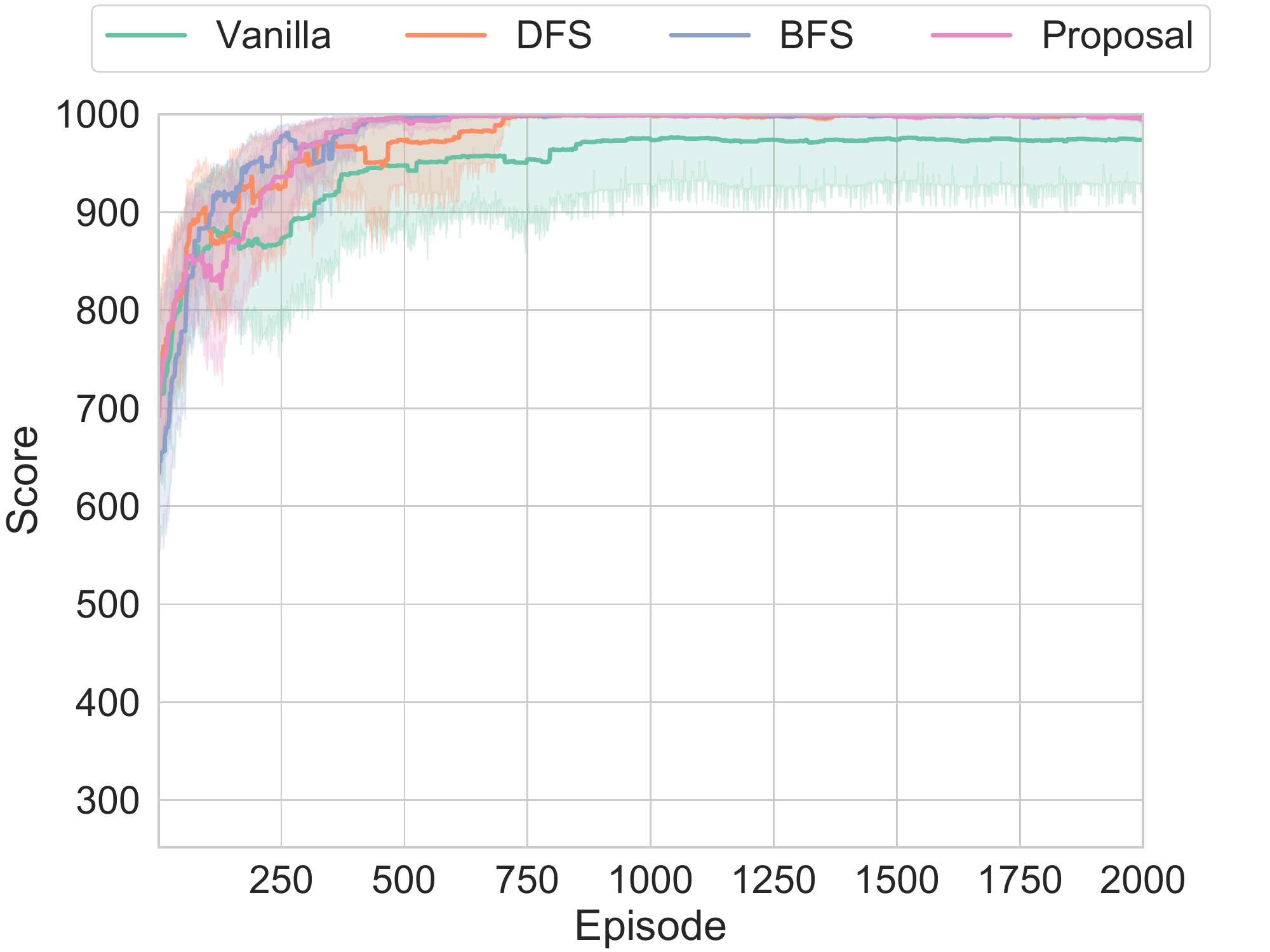}
        \subcaption{Reacher}
        \label{fig:result_score_Reacher}
    \end{subfigure}
    \begin{subfigure}[b]{0.32\linewidth}
        \centering
        \includegraphics[keepaspectratio=true,width=\linewidth]{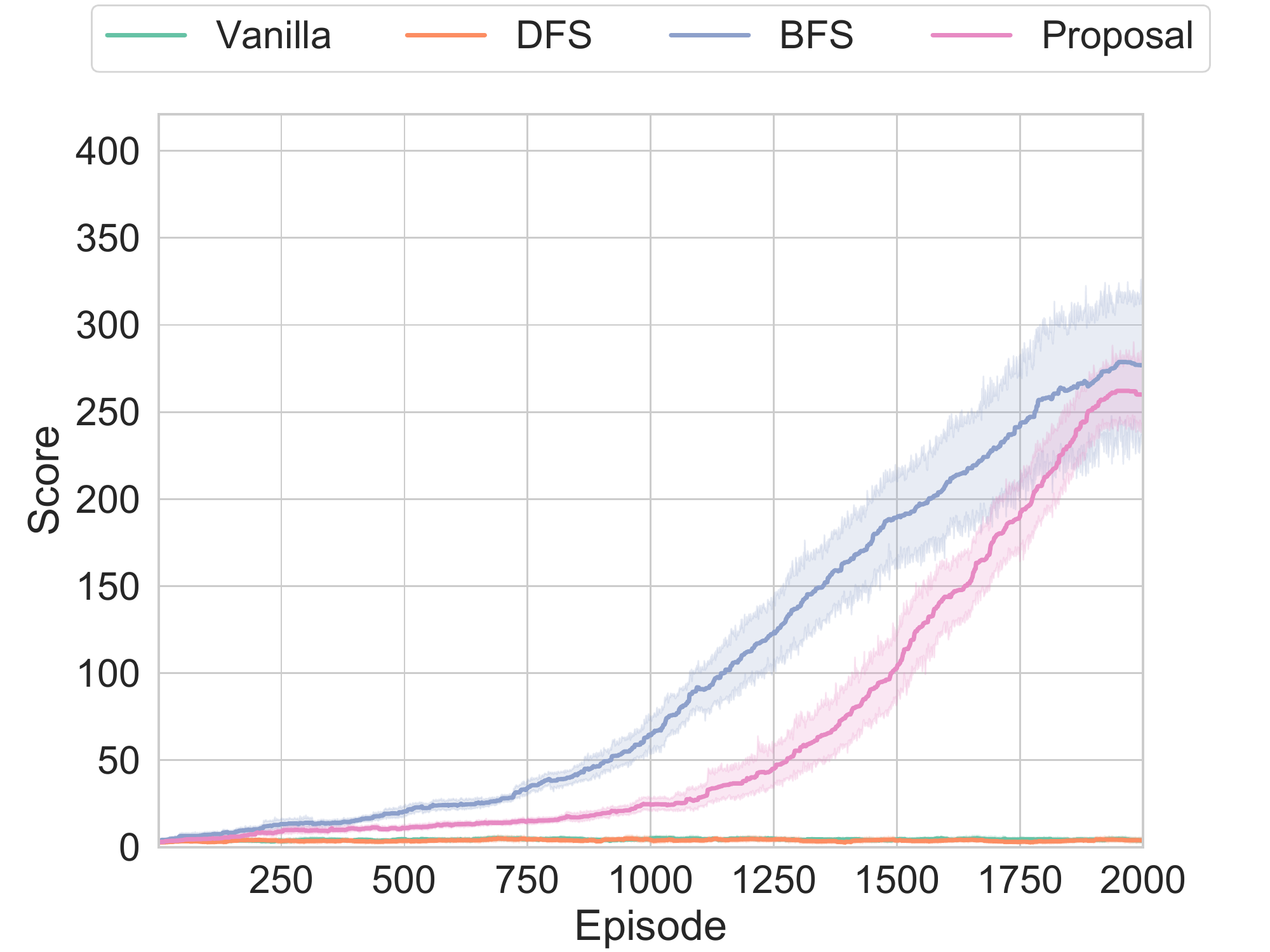}
        \subcaption{FingerSpin}
        \label{fig:result_score_FingerSpin}
    \end{subfigure}
    \caption{Learning curves}
    \label{fig:result_score}
\end{figure*}

\begin{table*}[tb]
    \caption{Test results: in each task, the best and second-best are in red and magenta texts, respectively.}
    \label{tab:result}
    \centering
    \begin{tabular}{l | c c c c c c}
        \hline\hline
        Method & \multicolumn{6}{c}{The sum of rewards (SD)}
        \\
               & Hopper & HalfCheetah & Ant & Cartpole & Reacher & FingerSpin
        \\
        \hline
        Vanilla
            & 1831 ($\pm$643)
            & 1432 ($\pm$555)
            & 2694 ($\pm$458)
            & 270 ($\pm$321)
            & 875 ($\pm$219)
            & 0 ($\pm$0)
        \\
        DFS
            & \textcolor{red}{1915} ($\pm$768)
            & 1711 ($\pm$639)
            & 2591 ($\pm$418)
            & \textcolor{magenta}{343} ($\pm$354)
            & \textcolor{magenta}{937} ($\pm$14)
            & 0 ($\pm$0)
        \\
        BFS
            & 1577 ($\pm$940)
            & \textcolor{magenta}{1799} ($\pm$580)
            & \textcolor{red}{2891} ($\pm$223)
            & 0 ($\pm$0)
            & 931 ($\pm$60)
            & \textcolor{magenta}{221} ($\pm$135)
        \\
        Proposal
            & \textcolor{magenta}{1912} ($\pm$724)
            & \textcolor{red}{1886} ($\pm$604)
            & \textcolor{magenta}{2865} ($\pm$249)
            & \textcolor{red}{453} ($\pm$358)
            & 917 ($\pm$121)
            & 187 ($\pm$111)
        \\
        \hline
        Mean
            & 1761 ($\pm$896)
            & 1703 ($\pm$615)
            & 2718 ($\pm$219)
            & 33 ($\pm$159)
            & 931 ($\pm$40)
            & 119 ($\pm$102)
        \\
        Sum
            & 1845 ($\pm$663)
            & 1706 ($\pm$555)
            & 2726 ($\pm$277)
            & 0 ($\pm$0)
            & \textcolor{red}{943} ($\pm$9)
            & \textcolor{red}{238} ($\pm$160)
        \\
        \hline\hline
    \end{tabular}
\end{table*}

\begin{figure}[tb]
    \centering
    \includegraphics[keepaspectratio=true,width=0.96\linewidth]{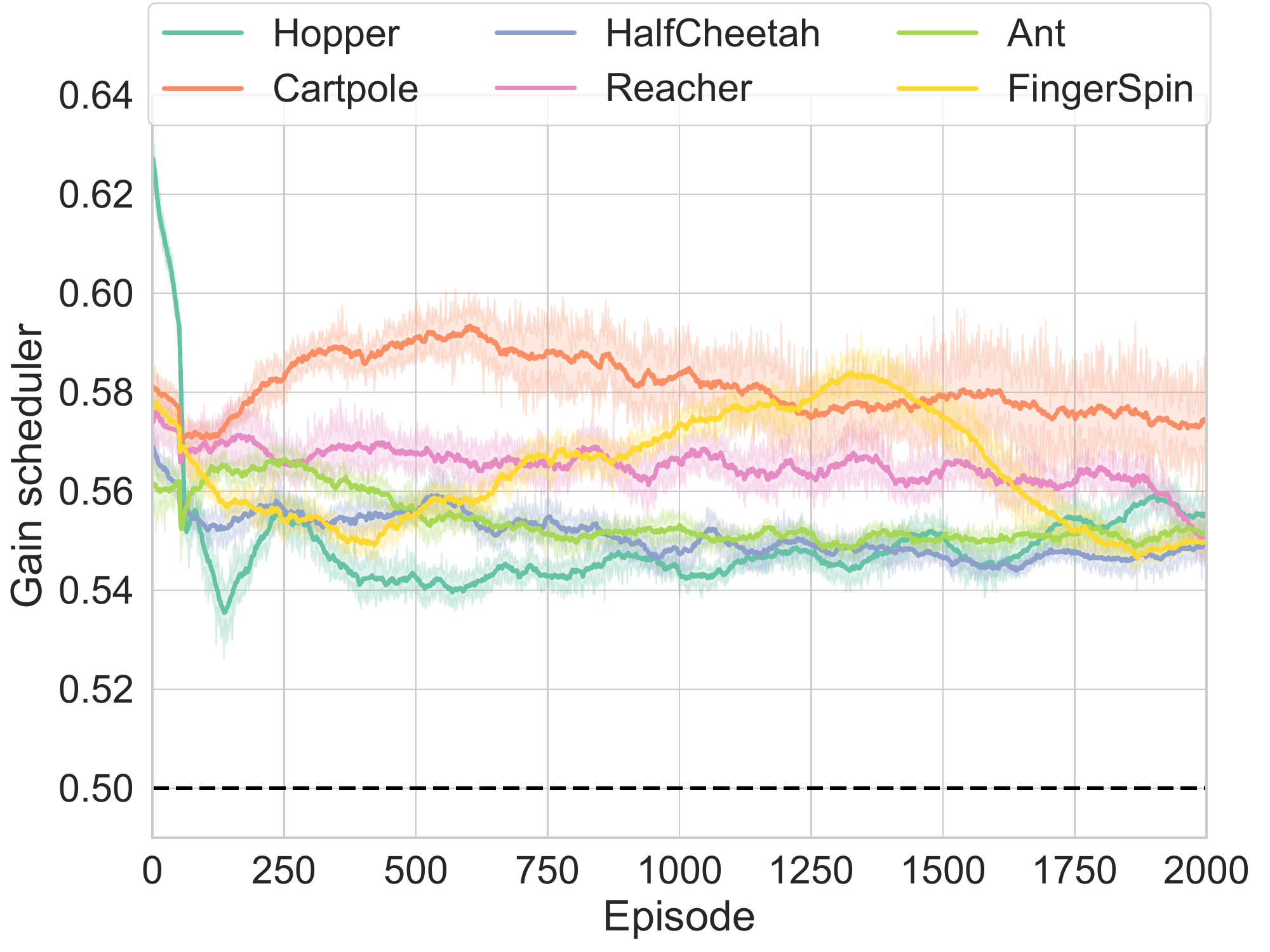}
    \caption{Adaptation of $\zeta$}
    \label{fig:result_gain}
\end{figure}

The learning curves for the four comparisons are depicted together in Fig.~\ref{fig:result_score}.
The results of the post-training tests, including the two ablation studies, are also summarized in Table~\ref{tab:result}.

At first, it is easily found that the DFS-like and BFS-like bonuses, defined in eqs.~\eqref{eq:bonus_dfs} and~\eqref{eq:bonus_bfs} respectively, contributes to improvement of learning performance compared to the case without them.
In particular, in the learning curves of HalfCheetah and Reacher, Vanilla seems to be stuck in local solutions, while the others succeeded in escaping them.
However, the tasks they excel at are different.
Namely, DFS failed to improve performance on Ant and FingerSpin and BFS completely failed on Cartpole.
In addition, BFS on Hopper achieved high learning efficiency at the beginning of learning, but eventually deteriorated its performance compared to the others, possibly due to over-learning.

In contrast, the proposed method, which appropriately combines these two bonuses with eq.~\eqref{eq:gain_schedule}, succeeded in all tasks by complementing each other's weak points.
As a result, the proposed method mostly included the top-2 methods in terms of the test results.
In addition, Mean with the fixed $\zeta=0.5$ was not in the top-2 methods in any task, indicating that the proposed gain scheduling itself worked effectively to adaptively alleviate the negative effects of each other's bonuses.
Even though, in the case of Reacher and FingerSpin, where the proposed method failed to be the top-2 methods, the best results were obtained by the simple sum of two bonuses, suggesting room for improvement (see the next discussion), although it hardly improved performance in the other tasks.

Finally, to confirm how the proposed method actually adjusted $\zeta$, $\zeta$ for each task is plotted in Fig.~\ref{fig:result_gain}.
As can be seen, $\zeta$ was adjusted to stay within $[0.5, 0.6]$ in each task, and mostly converged.
In addition, the convergence values on the respective tasks varied slightly.
For example, $\zeta$ was relatively larger for Cartpole, where DFS excelled, and $\zeta$ was relatively smaller for Ant, where BFS excelled.
Although the amount of variation would not be sufficient, the task-oriented switching between two bonuses was observed.
In addition, when $\zeta$ on FingerSpin is viewed with Fig.~\ref{fig:result_score_FingerSpin}, it can be found that before the score started to increase, the search in the depth direction was gradually advanced to get out of the region where no reward is given;
and after finding a way to do so, the search width was prioritized for a better solution.
As a result, although FingerSpin was able to find its optimal solution by simply increasing the search width, indicated in the results of BFS, the proposed method exerted the IDS-like search, as expected.
Note that the reason why $\zeta$ never converged to $0.5$ may be due to the limitation of optimization by the value range of $\kappa_{d,b}$ to be optimized for that purpose and the different variability of $m_{\kappa_{d,b}}$.

\subsection{Discussion}

From the above simulation results, it can be concluded that appropriately designed reward bonuses indeed assist learning and improve performance.
In addition, they have their own suitable tasks, especially the DFS- and BFS-like bonuses have complement with each other.
When combining them appropriately, therefore, learning performance for more versatile tasks can be expectedly improved, as like the proposed method in this paper.
However, there is still much room for improvements in the proposed method, two of the remaining issues are discussed as below.

One is the design of the individual bonuses and associated hyperparameters.
In particular, the value disagreement for DFS-like bonus may not be sufficient to explore unknown states deeper than the experienced states, since the value function is learned in a policy-dependent manner.
Utilizing the state transition model as in the previous study~\citep{pathak2019self} or extending it to introduce a world model~\citep{ha2018world} would be more appropriate for discriminating the faced situations, although it would increase the computational and learning costs.
For example, a variational autoencoder, which is frequently utilized in the world model, compresses the data used for training, i.e. the experienced states, into the center of latent space, which can be utilized as a discriminator for the unknown states~\citep{fujiishi2021safe}.
This may be effective in designing a new bonus that facilitate the search for the unknown states.
The increase in the computational and learning costs may not be a problem if the world model is effectively utilized in the context of model-based RL~\citep{chua2018deep,okada2020planet} as well.
However, any bonus design should require adjustment of hyperparameters, such as sensitivity, and should be used in conjunction with a meta-optimization method that efficiently adjusts them (e.g.~\citep{aotani2021meta}).

Another issue that revealed through the ablation studies is the superiority of a simple sum of two bonuses over the adaptive gain scheduling.
One of the reasons for this may be that the adaptive gain scheduling is basically optimized for $\zeta \simeq 0.5$, as shown in Fig.~\ref{fig:result_gain}, and does not allow either to dominate.
Of course, since which bonus is more effective for the given task is not clear, the equilibrium of $\zeta$ is reasonably $0.5$ at the beginning, but a mechanism to adjust it according to contribution of each bonus is desirable.
Although such a contribution is not easily estimated, the gradient information of value function that explicitly includes $\zeta$ as an input, for example, may be useful, as frequently employed in the context of explainable AI~\citep{das2020opportunities}.
Alternatively, while the proposed method uses $\lambda$ to define the overall bonus scale and distributes the usage rate among them by $\zeta \in [0, 1]$, an approach of establishing individual gains $\lambda_{d,b} \geq 0$ for the respective bonuses and optimizing them separately is also possible.
However, this approach has the risk of excessively large $\lambda_{d,b}$, which may not capture changes in the task-oriented reward.

\section{Conclusion}

This paper designed two different bonuses for RL based on the value disagreement and the self-imitation, respectively.
Noting that these bonuses are DFS- and BFS-like, properties of which are complement with each other, IDS-inspired adaptive gain scheduling is additionally designed to combine them appropriately.
That is, a heuristic first estimates whether the replayed experience is sufficiently learned and search is unnecessary.
If so, the search is deepened to find new situations, otherwise, the search is broadened to find the better states.
Numerical simulations confirmed that the designed DFS- and BFS-like bonuses are suitable for different tasks, and the proposed IDS-inspired method inherited their properties and successfully learned on all tasks.

However, as discussed, there is still room for improvement in the heuristic designs, which will be resolved in the near future.
In addition, benefits of improved search efficiency will be further investigated through applications such as motion control of autonomous robots in real world.

\appendix
\section{Implementation of baseline RL}
\label{app:impl}

\begin{table*}[tb]
    \caption{Configuration of the baseline RL}
    \label{tab:base}
    \centering
    \begin{tabular}{ccc}
        \hline\hline
        Symbol & Meaning & Value
        \\
        \hline
        $(N, L)$ & \#Neuron and \#hidden layer & $(100, 2)$
        \\
        --- & Activation function & LayerNorm~\citep{ba2016layer} + Squish~\citep{barron2021squareplus}
        \\
        $\gamma$ & Discount factor & $0.99$
        \\
        $(K, \beta)$ & For ensemble learning & $(10, 1)$
        \\
        $(\kappa, \beta, \lambda, \underline{\Delta})$ & For PPO-RPE~\citep{kobayashi2022proximal} & $(0.5, 0.5, 0.999, 0.1)$
        \\
        $(N_c, N_b, N_r, \alpha, \beta)$ & For PER~\citep{schaul2015prioritized} & $(12800, 32, 6400, 1.0, 0.5)$
        \\
        $(\sigma, \underline{\lambda}, \overline{\lambda}, \beta)$ & For L2C2~\citep{kobayashi2022l2c2} & $(1.0, 0.01, 1.0, 0.1)$
        \\
        $(\tau, \underline{\tilde{\nu}}, \lambda, q)$ & For CAT-soft update~\citep{kobayashi2022consolidated} & $(0.1, 1, 1, 1)$
        \\
        $(\alpha, \beta, \epsilon, \underline{\tilde{\nu}})$ & For AdaTerm~\citep{ilboudo2022adaterm} & $(10^{-3}, 0.9, 10^{-5}, 1.0)$
        \\
        \hline\hline
    \end{tabular}
\end{table*}

RL used in this paper is implemented with PyTorch~\cite{paszke2017automatic} and the same as the one used in the literature~\citep{kobayashi2022consolidated}.
Specifically, two-layered fully connected networks with 100 neurons for each are assigned as the hidden layers to approximate the policy and value functions independently.
The activation function for them is with LayerNorm~\citep{ba2016layer} and Squish, which is given as $x \sigma_s(x)$ with $\sigma_s$ derivative of Squareplus~\cite{barron2021squareplus}.
The stochastic policy function is modeled by student-t distribution~\citep{kobayashi2019student} for efficient search.
As discussed in the main text, the value function is approximated by the ensemble structure~\cite{osband2016deep} (see \ref{app:median}).

Under such a network architecture, the learning algorithm is implemented as below.
Note that the hyperparameters used are listed in Table~\ref{tab:base}, but if no explanation is given, the definition of each variable remains the same as in the original paper, duplicating assignment of several variables.
The basis of the learning algorithm is PPO-RPE~\citep{kobayashi2022proximal}, which allows smooth update of the policy.
The limited update amount of the policy allows the use of experience replay even in on-policy RL, and therefore, PER~\citep{schaul2015prioritized} is introduced with a capacity of $N_c$ to replay up to $N_r$ experiences at the end of each episode per $N_b$ batch size.
To suppress over-learning of function approximation, regularization by L2C2~\citep{kobayashi2022l2c2} is given as an auxiliary problem, restricting the shape of the policy and value functions to be smooth locally.
The pseudo-supervised signals in TD learning (and the actions for getting experiences) are generated from the target networks, and CAT-soft update~\citep{kobayashi2022consolidated} is employed for updating them to prevent learning delay and instability.
Finally, the optimization parameters, such as the network weights and biases, are updated with AdaTerm~\cite{ilboudo2022adaterm}, a type of stochastic gradient descent method that is robust to noise in the pseudo-supervised signals.

\section{Median operation for ensemble learning of value function}
\label{app:median}

\begin{table}[tb]
    \caption{Evaluation of the median operation}
    \label{tab:median}
    \centering
    \begin{tabular}{l c | c}
        \hline\hline
        Method & $(K, \beta)$ & The sum of rewards (SD)
        \\
        \hline
        No ensemble & $(1, 0)$ & 6220 ($\pm$3056)
        \\
        Mean        & $(10, 0)$ & 6660 ($\pm$3415)
        \\
                    & $(10, 1)$ & 5823 ($\pm$3252)
        \\
        Median      & $(10, 0)$ & 7715 ($\pm$3002)
        \\
                    & $(10, 1)$ & 8377 ($\pm$2272)
        \\
        \hline\hline
    \end{tabular}
\end{table}

The consensus operations for the ensemble of value functions $V$ are compared.
The benchmark is \textit{InvertedDoublePendulumBulletEnv-v0} implemented on Pybullet~\citep{coumans2016pybullet} with white noise (scale is $0.001$), and is evaluated by the test results after training using each method 24 times with different random seeds as in the above simulations.
The algorithm is identical to the main simulations, although the proposed bonuses are omitted for brevity.

The sum of rewards after training are shown in Table~\ref{tab:median}.
The mean operation hardly improved the performance from no ensemble case.
On the contrary, adding the prior degraded the performance.
This may be due to the fact that the increase in variance caused by the prior in a small number of ensembles caused a sample bias, which destabilized learning.

In contrast, the median operation achieved a significant improvement in performance with and without the prior.
Especially with the prior, the task was successful to some extent even in the worst case, as can be seen in the decreased standard deviation.
Therefore, it can be concluded that the median operation with the prior stabilizes RL, although the optimal $K$ and $\beta$ are under discussion.

\section*{Acknowledgments}

This work was supported by JSPS KAKENHI, Grant-in-Aid for Scientific Research (B), Grant Number JP20H04265.

\bibliographystyle{elsarticle-harv}
\bibliography{biblio}

\end{document}